\documentclass{article}

     \PassOptionsToPackage{numbers, compress}{natbib}

 \usepackage[final]{neurips_2019}

\usepackage[utf8]{inputenc} 
\usepackage[T1]{fontenc}    
\usepackage{hyperref}       
\usepackage{url}            
\usepackage{booktabs}       
\usepackage{amsfonts}       
\usepackage{nicefrac}       
\usepackage{microtype}      

\usepackage{array, booktabs, makecell}
\usepackage{graphicx}
\usepackage{amsmath}

\title{Learning Embeddings from Cancer Mutation Sets for Classification Tasks}

\makeatletter
\newcommand{\printfnsymbol}[1]{%
  \textsuperscript{\@fnsymbol{#1}}%
}
\makeatother

\author{
  Geoffroy Dubourg-Felonneau$^1$, Yasmeen Kussad$^{1,2}$, Dominic Kirkham$^{1,3}$, \\
  \textbf{John W Cassidy$^1$, Nirmesh Patel$^1$, Harry W Clifford$^1$} \\
  \\
  \\
  \and
  $^1$Cambridge Cancer Genomics \\
  Cambridge, UK \\
  www.ccg.ai \\
  \and
  $^2$University of Lancaster \\
  Lancaster, UK \\
  \and
  $^3$University of Cambridge \\
  Cambridge, UK \\
}
\begin{document}

\maketitle

\begin{abstract}

Analysis of somatic mutation profiles from cancer patients is essential in the development of cancer research. However, the low frequency of most mutations and the varying rates of mutations across patients makes the data extremely challenging to statistically analyze as well as difficult to use in classification problems, for clustering, visualization or for learning useful information. Thus, the creation of low dimensional representations of somatic mutation profiles that hold useful information about the DNA of cancer cells will facilitate the use of such data in applications that will progress precision medicine. In this paper, we talk about the open problem of learning from somatic mutations, and present Flatsomatic: a solution that utilizes variational autoencoders (VAEs) to create latent representations of somatic profiles. The work done in this paper shows great potential for this method, with the VAE embeddings performing better than PCA for a clustering task, and performing equally well to the raw high dimensional data for a classification task. We believe the methods presented herein can be of great value in future research and in bringing data-driven models into precision oncology.

\end{abstract}

\section{Introduction}

As part of the decision-making process in cancer therapy selection, oncologists stratify patients into broad clinical groups. At present, the major indicator for stratification is the site of origin of the primary tumor. However, with the widespread adoption of next generation sequencing technologies, our reference for stratification is becoming more complex as we are beginning to understand how each tumor is unique on the genetic level. The outcome of which is an approach dubbed precision oncology, which involves the process of identifying genomic features driving an individual tumor and designing a personalized therapeutic strategy in response.

This presents a classification problem that is well suited to supervised machine learning algorithms, although high complexity, sparisty, varying set sizes, and high dimensionality of genomic data such as somatic mutations, makes this difficult. Models cannot be applied directly on raw somatic mutation data, and sub-optimal data transformation of these data into something suitable will yield sub-optimal performance of the resultant model.

In this paper we discuss the open problem of learning from somatic mutations and we present a solution, Flatsomatic: a Variational Auto Encoder optimized to compress somatic mutations to allow for unbiased data compression whilst maintaining  signal. We hope the methods presented here can enable greater use of somatic mutation information in machine learning algorithms.

\section{The Data}

Somatic mutation input data for a given patient is a set of mutations $\{M_1, ..., M_n\}$ with $n$ representing the number of mutations. Somatic mutation data $M_i$ has features including (but not limited to) chromosome number (chr), position in chromosome (pos), and Variant Allele Frequency (VAF). For The Cancer Genome Atlas (TCGA)\cite{RN30} dataset, we observe these features across 8062 patients, with anywhere between 2 and 1000 mutations from a set of millions of possible mutations.

Our approach to these data is to limit ourselves to positional features only (chromosome number, position in chromosome) and to reduce the mutations occurring in a given location a same datum by applying a surjection that concatenates the positional features. 

\begin{figure}[h!]
  \centering
  \includegraphics[width=0.4\linewidth]{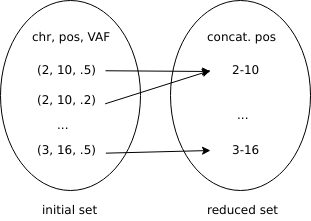}
  \caption{Application of a concatenation function on a subset of features from the initial set (left) creates a set (right) of lower cardinality}
  \label{figure:vae-training}
  \end{figure}

The advantage of this method is a decrease in space to 136000 possible mutations and the creation of a matrix of mutation occurrences from which it is easy to train:

\[
mutation\_occurrences
=
\begin{bmatrix}
    O_{11} & \dots  & O_{1m} \\
    \vdots & \ddots & \vdots \\
    O_{n1} & \dots  & O_{nm} \\
\end{bmatrix}
\]

$n$ : number of samples \\
$m$ : number of observed mutations \\
$O_{ij}$ : 1 if mutation $j$ occurs in sample $i$, 0 otherwise.

Finally, as such data is challenging to statistically analyze and difficult to use in classification/clustering/visualisation, the creation of low dimensional representations of somatic mutation profiles is necessary. These hold useful information about the DNA of cancer cells and will facilitate the use of such data in downstream analysis.

\section{Methods}
 We apply variational autoencoders (VAEs) \cite{RN27} to create latent representations of somatic profiles from the $mutation\_occurrences$ matrix. The process of building the VAEs to compress somatic profiles included implementation and optimization of several different neural network architectures. Several changes to the loss function of the VAE were explored \cite{RN15} in order to learn more useful representations. All models were implemented using the Keras library \cite{RN40} with TensorFlow backend. 

The somatic profiles used to train our models were comprised of the aforementioned 8062 TCGA pan-cancer patients, and 989 cell lines from the COSMIC cell line project (CCLP). The profiles for each patient are represented by the mutation position in the genome. To pre-process the data, binary vectors of equal lengths (all mutation locations) were created for each patient. To reduce the sparsity, the mutation locations with a frequency of less than 5 were removed. To assess the reconstruction ability of the VAE, we performed a 5 fold cross-validation and used the micro F1-score that we will simply refer as F1-score. 
\subsection{Multi Layer Perceptron based models}
The first architecture explored was a feed-forward network also known as multilayer perceptron (MLP). A two-layer deep network in the encoder/decoder of the VAE were built with a batch normalization \cite{RN39} layer after each layer. Different combinations of the number of units in each layer were attempted and the effect of changing the size of the latent space was studied. A leaky ReLU \cite{LRELU} activation was used in the encoder layers, and a regular ReLU was used in the decoder layers except for the final layer in the decoder which employed a sigmoid function. A dropout layer was added after the first layer in the encoder/decoder, and  L1 regularization was used with each dense layer. We optimized with RMSprop, and the models were trained with a batch size of 128 for 100 epochs.

\subsection{Bi-directional LSTM}
The second architecture attempted was a bidirectional LSTM \cite{Schuster1997BidirectionalRN} to reduce the significance of the order of mutations, especially those toward the end of the input sequence. Although the order of mutations in a somatic profile is not important, we expect the occurrence of certain mutations or a group of mutations together to hold important information, hence this architecture was deemed suitable to learn deeper patterns in the data. The number of units in the LSTM was explored (1024, 512, 256, or 128) as well as the size of the latent space (128, 64, 32, or 8).

\subsection{Changes in Loss Function of VAE}
The loss function of the VAE is comprised of a reconstruction loss and a variational regularization term. The loss function is essential to help the decoder reconstruct a version of the data from the embeddings as close as possible to the input data. Binary cross-entropy is frequently used as the reconstruction loss, however, we devised an F1-score based loss function to use as the reconstruction loss and compared its performance to binary cross-entropy. 

\[L=\mathbb{E}_{z\sim q}[\log p(x\mid z)]-D_{KL}(q(z\mid x)\mid \mid p(z))\]

Recent studies \cite{RN33} have emphasized the importance of the regularization term in the loss function and the role it plays in creating useful latent representations. To address this, we explored the use of a beta-VAE \cite{RN34}.

\subsection{Assessment of Representation}
To assess the representation created by the VAE, 2-dimensional embeddings were created by the best VAE and compared to a lower dimension version of the data created using principal component analysis (PCA) \cite{RN16}. Kmeans clustering was then applied to both low-dimensional versions, and results were compared to the known cancer types of each profile using Normalized Mututal Information (NMI) as a measure. The number of clusters created was 32, matching the number of cancer types in these data. The embeddings were also used in a given classification task and their performance was compared to results when the raw data was used in the same task.

\section{Results}

Initial results show the MLP-VAE provides a better performance than the bidirectional LSTM VAE, with F1-scores  of 20.4\%, and 17.1\% respectively. Our analysis shows a latent space size of >8 is required for optimum reconstruction ability of the VAE. Table \ref{table:latent-spaces} shows the validation F1, found with different latent space sizes.



\begin{table}[h!]
  \caption{The effect of Latent Space Size on VAE reconstruction}
  \label{table:latent-spaces}
  \centering
  \begin{tabular}{cccccccc}
    \toprule
    \textbf{Latent Space Size} & 2 & 8 & 32 & 64 & 128 & 265 & 512 \\
    \midrule
    \textbf{F1-score} & 15.68 & 20.55 & 20.88 & 20.41 & 20.50 & 20.18 & 20.21\\
    \bottomrule
  \end{tabular}
\end{table}

We found that the use of F1-score based loss helps the VAE reconstruct this form of data much better than the binary cross-entropy loss. To assess this, we measure the cosine similarity. Figure \ref{figure:vae-training} shows the performance of the VAE while training with the two losses. We also found that introducing the regularization term using a warm-up function and increasing its weight with each epoch gives the best performance.

\begin{figure}[h!]
  \centering
  \includegraphics[width=0.7    \linewidth]{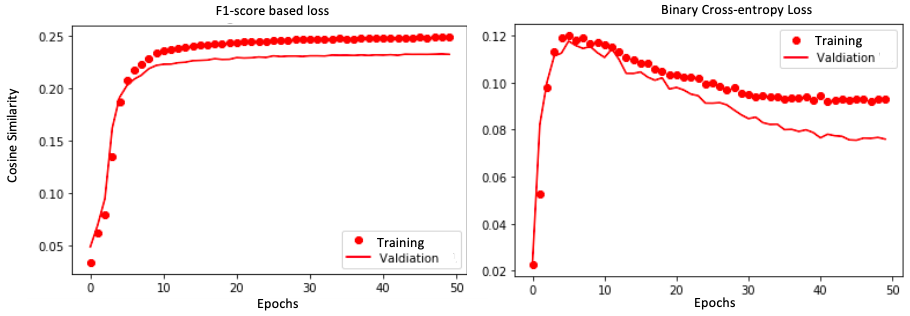}
  \caption{Comparison of F1-loss (left) and Cross-entropy (right) during training}
  \label{figure:vae-training}
\end{figure}
  
\subsection{Visualization}

After performing K-means on the embeddings from the best VAE and the lower dimension of the data created by PCA, plots with reference to the clusters were created. The NMI for the clusters created in the VAE embeddings was 21\%, and for the version created by PCA it was 11\%, showing that the VAE embeddings are a better representation.


\subsection{Use-case}

We have used Flatsomatic in a semi-supervised setting to predict drug response from the GDSC \cite{yang_genomics_2013} dataset. We show (Table \ref{table:drug-response}) that the performances are similar to the raw data. The advantage of working with smaller spaces is that it enables the use of other relevant features (either clinical features or other abstract representations of genomic profiles) for future work.

\begin{table}[h!]
  \caption{Predicting drug response with somatic profile}
  \label{table:drug-response}
  \centering
  \begin{tabular}{ccccc}
    \toprule
    Data source & Dimensionality & Precision & Recall & \textbf{F1-score} \\
    \midrule
    Mutation occurrences  & 8298 & 0.732 & 0.614 & 0.667 \\
    Flatsomatic embeddings & 64 & 0.721 & 0.621 & 0.667 \\
    \bottomrule
  \end{tabular}
\end{table}

\section{An Open Problem}

For future work we would like to investigate better ways of encoding the initial set of mutations through retaining the non-positional features and through not introducing sparsity by creating a $mutation\_occurrences$ matrix. We believe that the pre-processing step that consists in reducing the data into matrices of mutation occurrence is sub-optimal as non-positional features are discarded and it is assumed that two mutations at the same position will have the same effect.

We also aim to explore additional methods for encoding such data for machine-learning purposes. One could, for example, encode additional mutation data with a vector including non-positional features (Variant Allele Frequency, Impact, Missense Nonsense, Base change, etc) and implement a model that can utilize such variable length inputs, such as a Recurrent Neural Network.

\section{Conclusion}

The work done in this paper has shown that there is a great potential for the use of VAEs in creating utilizable lower dimension representations of somatic profiles. The VAE embeddings performed better than PCA for a clustering task, and performed equally well to the raw high dimensional data for a classification task. We believe that this work is an important step in the use of somatic mutation data for machine learning applications, and will help future researchers in bringing data-driven models into precision oncology.




\bibliography{neurips_2019.bib}

\begin{thebibliography}{10}

\bibitem{RN30}
Network The Cancer Genome Atlas~Research et~al.
\newblock The cancer genome atlas pan-cancer analysis project.
\newblock {\em Nature Genetics}, 45:1113, 2013.

\bibitem{RN27}
Y.~Bengio, A.~Courville, and P.~Vincent.
\newblock Representation learning: A review and new perspectives.
\newblock {\em IEEE Transactions on Pattern Analysis and Machine Intelligence},
  35(8):1798--1828, 2013.

\bibitem{RN15}
Diederik~P. Kingma and Max Welling.
\newblock {\em Auto-Encoding Variational Bayes}, volume abs/1312.6114.
\newblock 2013.

\bibitem{RN40}
Francois Chollet et~al.
\newblock Keras.
\newblock 2015.

\bibitem{RN39}
Sergey Ioffe and Christian Szegedy.
\newblock Batch normalization: Accelerating deep network training by reducing
  internal covariate shift.
\newblock {\em arXiv preprint arXiv:1502.03167}, 2015.

\bibitem{LRELU}
Kaiming He, Xiangyu Zhang, Shaoqing Ren, and Jian Sun.
\newblock Delving deep into rectifiers: Surpassing human-level performance on
  imagenet classification, 2015.

\bibitem{Schuster1997BidirectionalRN}
Mike Schuster and Kuldip~K. Paliwal.
\newblock Bidirectional recurrent neural networks.
\newblock {\em IEEE Trans. Signal Processing}, 45:2673--2681, 1997.

\bibitem{RN33}
Alexander~A Alemi, Ben Poole, Ian Fischer, Joshua~V Dillon, Rif~A Saurous, and
  Kevin Murphy.
\newblock Fixing a broken elbo.
\newblock {\em arXiv preprint arXiv:1711.00464}, 2017.

\bibitem{RN34}
Irina Higgins, Lo'efc Matthey, Arka Pal, Christopher Burgess, Xavier Glorot,
  Matthew~M Botvinick, Shakir Mohamed, and Alexander Lerchner.
\newblock {\em beta-VAE: Learning Basic Visual Concepts with a Constrained
  Variational Framework}.
\newblock ICLR. 2017.

\bibitem{RN16}
Svante Wold, Kim Esbensen, and Paul Geladi.
\newblock Principal component analysis.
\newblock {\em Chemometrics and Intelligent Laboratory Systems}, 2(1):37--52,
  1987.

\bibitem{yang_genomics_2013}
W.~Yang, J.~Soares, P.~Greninger, E.~J. Edelman, H.~Lightfoot, S.~Forbes,
  N.~Bindal, D.~Beare, J.~A. Smith, I.~R. Thompson, S.~Ramaswamy, P.~A.
  Futreal, D.~A. Haber, M.~R. Stratton, C.~Benes, U.~McDermott, and M.~J.
  Garnett.
\newblock Genomics of {{Drug Sensitivity}} in {{Cancer}} ({{GDSC}}): A resource
  for therapeutic biomarker discovery in cancer cells.
\newblock 41:D955--61.

\end{thebibliography}
\bibliographystyle{unsrt}

\medskip

\end{document}